\documentclass{article}

\usepackage[dblblindworkshop, final]{neurips_2025}

\usepackage[utf8]{inputenc} 
\usepackage[T1]{fontenc}    
\usepackage{hyperref}       
\usepackage{url}            
\usepackage{booktabs}       
\usepackage{amsfonts}       
\usepackage{nicefrac}       
\usepackage{microtype}      

\hypersetup{hidelinks}

\usepackage{graphicx}
\usepackage{xcolor}
\usepackage{soul}
\newcommand{\highlight}[1]{\colorbox{yellow}{#1}}
\usepackage{adjustbox}

\usepackage{bidi}          

\usepackage{fontspec}

\newfontfamily\farifont{NotoSansArabic-Regular.ttf}
\newfontfamily\hindifont{NotoSansDevanagari-Regular.ttf}
\newfontfamily\mandarinfont{NotoSansCJKsc-Regular.otf}

\title{Investigating Hallucination in Conversations for Low Resource Languages}

\author{
 \textbf{Amit Das\textsuperscript{1}},
 \textbf{Md. Najib Hasan\textsuperscript{2}},
 \textbf{Souvika Sarkar\textsuperscript{2}},
 \textbf{Zheng Zhang\textsuperscript{3}}, \\
 \textbf{Fatemeh Jamshidi\textsuperscript{4}},
 \textbf{Tathagata Bhattacharya\textsuperscript{5}},
 \textbf{Nilanjana Raychawdhary\textsuperscript{6}},
 \\
 \textbf{Dongji Feng\textsuperscript{7}},
 \textbf{Vinija Jain\textsuperscript{8,9\,\textasteriskcentered}},
 \textbf{Aman Chadha\textsuperscript{8, 10}}\thanks{Work does not relate to position at Amazon or Meta.}
\\
\\
 \textsuperscript{1}University of North Alabama,
  \textsuperscript{2}Wichita State University,
 \textsuperscript{3}Murray State University, \\ 
 \textsuperscript{4}California State Polytechnic University Pomona,
 \textsuperscript{5}Auburn University at Montgomery, \\ 
\textsuperscript{6}Auburn University,
 \textsuperscript{7}Gustavus Adolphus College,
 \textsuperscript{8}Stanford University, \\ 
 \textsuperscript{9}Meta,
 \textsuperscript{10}Amazon GenAI 
\\
\small{
  \textbf{Corresponding author:} Amit Das (\href{mailto:email@domain}{adas@una.edu})
 }
}

\begin{document}

\maketitle

\begin{abstract}
Large Language Models (LLMs) have demonstrated remarkable proficiency in generating text that closely resemble human writing. However, they often generate factually incorrect statements, a problem typically referred to as `hallucination'. Addressing hallucination is crucial for enhancing the reliability and effectiveness of LLMs. While much research has focused on hallucinations in English, our study extends this investigation to conversational data in three languages: Hindi, Farsi, and Mandarin. We offer a comprehensive analysis of a dataset to examine both factual and linguistic errors in these languages for GPT-3.5, GPT-4o, Llama-3.1, Gemma-2.0, DeepSeek-R1 and Qwen-3. We found that LLMs produce very few hallucinated responses in Mandarin but generate a significantly higher number of hallucinations in Hindi and Farsi.
\end{abstract}

\section{Introduction}

LLMs have emerged as a groundbreaking advancement in artificial intelligence, revolutionizing the field of natural language processing (NLP). These models show an extraordinary ability to perform a wide range of language-related tasks such as text generation, machine translation, summarization, question answering, etc. Despite their impressive performance, a critical challenge persists: the phenomenon of `hallucination,' where LLMs produce text that is factually incorrect, nonsensical, or misleading \cite{ji2023survey, li2023halueval}. This issue is not limited to English but also affects low-resource languages.

Hallucination in neural language generation has become a growing concern across various areas, including neural machine translation \cite{raunak2021curious, wang2020exposure} and summarization \cite{durmus2020feqa, kang2020improved}. Research on LLMs has highlighted problems like generating inaccurate information, producing incoherent text, and merging unrelated sources \cite{mckenna2023sources}. In the context of machine translation, studies have reported hallucination rates exceeding 10\% for certain language pairs \cite{guerreiro2023hallucinations}. Despite these advancements, the issue of hallucination in conversational systems, especially in low-resource languages, remains largely underexplored.

The complexity and significance of low-resource languages underscore the urgent need for a detailed investigation into hallucinations within these linguistic contexts. Addressing this issue not only improves the reliability and applicability of LLMs but also has implications for a range of applications, including information retrieval, sentiment analysis, and machine translation. This study aims to rigorously evaluate the factual accuracy and reliability of LLMs, specifically GPT-3.5 and GPT-4o, when generating texts in Hindi, Farsi, and Mandarin.

\begin{figure*}[h!]
    \includegraphics[scale=.41]
    {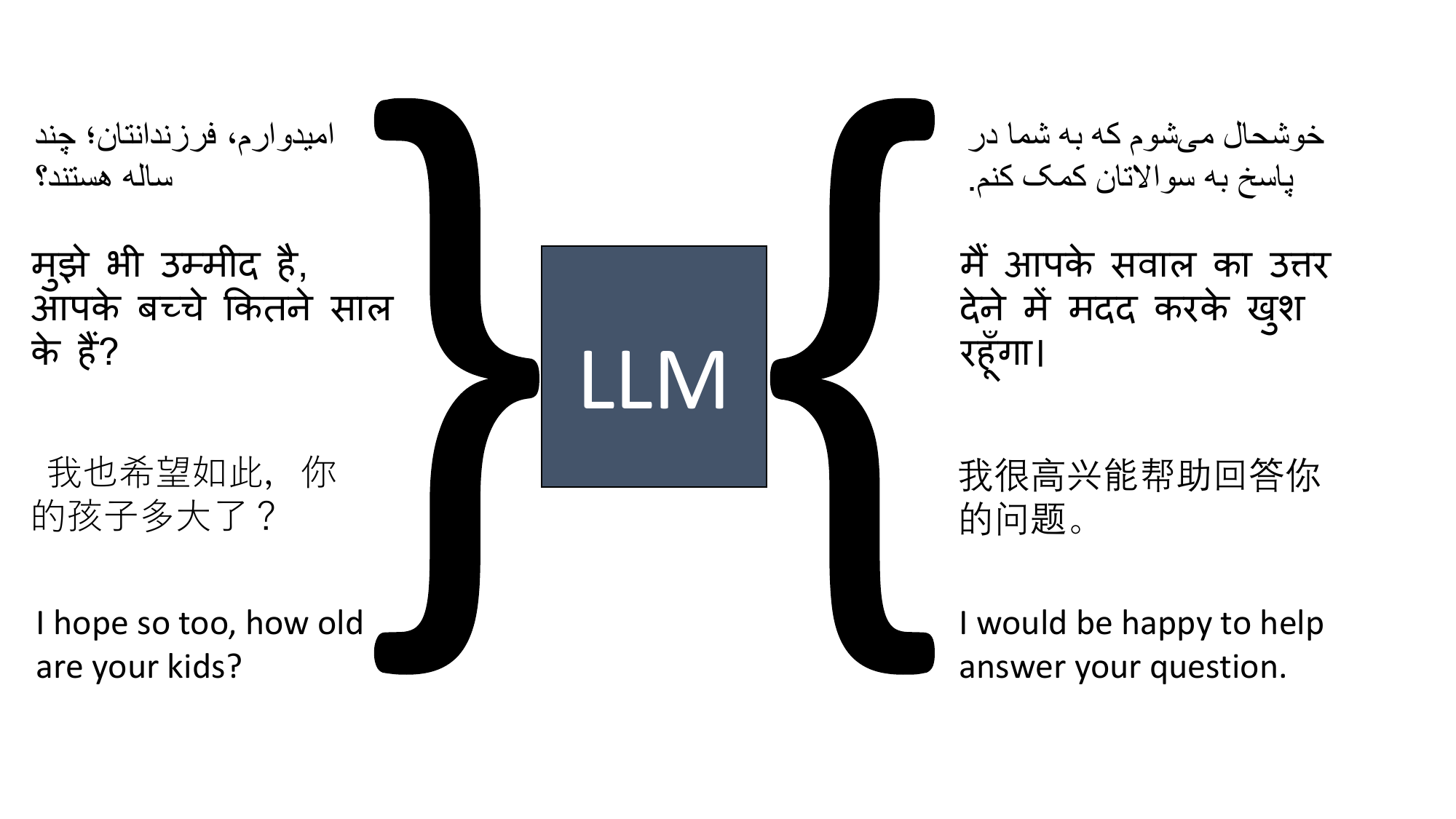}
    \centering
    \caption{Workflow diagram of our work. It shows a sample conversation where an LLM provides irrelevant response for Hindi, Farsi and Mandarin. The left side are the inputs to the LLM and right side are the irrelevant responses by LLM. We have explored GPT-3.5, GPT-4o, Llama-3.1, Gemma-2.0, DeepSeek-R1 and Qwen-3 in this paper.}
    \label{fig:workflow}
\end{figure*}

Our research seeks to make the following contributions:

1. We conduct an analysis of the factual accuracy of outputs of LLMs (specifically for GPT-3.5, GPT-4o, Llama-3.1, Gemma-2.0, DeepSeek-R1 and Qwen-3) in Hindi, Farsi, and Mandarin, offering insights into their reliability across these languages.\footnote{Sample LLM responses can be found here: \url{https://anonymous.4open.science/r/LLM-Hallucination-Low-Resource-Languages-6C98/}. We will provide all the relevant codes and responses if the paper is accepted.}

2. We identify and categorize the types of factual and linguistic errors present in conversational outputs generated by LLMs for Hindi, Farsi, and Mandarin.

\section{Related Work}

LLMs are increasingly scrutinized for both their susceptibility to bias and their tendency to hallucinate, with recent research revealing concerning patterns across diverse NLP tasks. Various forms of bias have been identified—including gender, religion, and political ideology—affecting both model behavior and output generation \cite{salinas2023not, jenny2023navigating, he2023inducing}. These biases are not only pervasive but also task-sensitive, as shown by Zheng et al. \cite{zheng2023large}, who demonstrate that simple perturbations such as altering the order of multiple-choice options can lead to selection bias. Similarly, Gonçalves and Strubell \cite{gonccalves2023understanding} offer a comprehensive synthesis of how biases manifest in LLMs, while others have proposed mitigation techniques by retraining on bias-filtered datasets \cite{dong2023co, limisiewicz2023debiasing}. However, as highlighted in our Section 3, such mitigation techniques may not generalize well, particularly in complex tasks like stance detection, where the bias distribution itself varies significantly.

Hallucination remains a central challenge in LLM deployment \cite{dale2022detecting}. Initial work in this area focused mainly on grounded tasks such as summarization and dialogue, where hallucinations were detected by aligning the generated factual units with the input of the source using entailment-based or QA-driven techniques \cite{maynez2020faithfulness}. As hallucination detection has expanded to open source generation, newer approaches now use reliable references such as Wikipedia and web searches to assess factuality \cite{min2023factscore, chern2023factool, mishra2024fine}, while others focus on specific domains such as citation verification \cite{agrawal2023language}. Complementing these reference-based methods, internal LLM techniques, such as consistency checks across sampled responses \cite{manakul2023selfcheckgpt} and logit-space analysis \cite{varshney2023stitch}, aim to identify contradictions without external validation.

Together, these lines of work underscore the intertwined nature of bias and hallucination in LLMs. Both phenomena reflect deeper model limitations in generalization, factual grounding, and representation learning—issues that continue to motivate new strategies for detection and mitigation.

\section{Methodology}

\begin{table*}[h!]
\centering
\caption{Sample translations of conversations from English into Hindi, Farsi and Mandarin using GPT-3.5. A sample prompt we used for translating the texts to Farsi is: \texttt{'Translate the following into Farsi: '}. Each translation was then manually checked to ensure translation quality.}
\vspace{\baselineskip}
\begin{tabular}{p{2.9cm} p{2.9cm} p{2.9cm} p{2.9cm}}
\toprule
\textbf{English} & \textbf{Farsi} & \textbf{Hindi} & \textbf{Mandarin} \\
\midrule
1: `Which level are you at?' \newline 2: `I received on-the-job training when I first started' &
\RL{1: \farifont  «در چه سطحی هستید؟»} \newline 2: \RL{\farifont «وقتی اولین بار شروع کردم، آموزش‌های حین کار دریافت کردم.»} &
1: {\hindifont 'आप किस स्तर पर हैं?'} \newline 2: {\hindifont 'जब मैंने पहली बार शुरू किया था, तो मुझे ऑन-द-जॉब प्रशिक्षण मिला।'} & 1:
{\mandarinfont 你现在处于哪个水\newline 平？} \newline 2: {\mandarinfont 我刚开始时接受了 \newline 在职培训。} \\
1: `That's great! How long have you been doing this work?' \newline 2: `For a good number of years now.' &
\RL{1: \farifont  «عالیه! چقدر مدت است که این کار را انجام می‌دهید؟»} \newline 2: \RL{\farifont «برای چندین سال خوب.»} &
1:{\hindifont  'यह बहुत अच्छा है! आप यह काम कितने समय से कर रहे हैं?'} \newline 2:{\hindifont 'काफी सालों से।'} &
1: {\mandarinfont  太好了！你做这份 \newline 工作的时间有多久 \newline 了？} \newline 2: {\mandarinfont 已经有好几年了。} \\
1: `Sounds like it could be a dangerous job at times.' \newline 2: `That it is, especially if you don't take the proper measures' &
\RL{1:\farifont  «به نظر می‌رسد گاهی اوقات می‌تواند شغل خطرناکی باشد.»} \newline 2: \RL{\farifont «همینطور است، به خصوص اگر اقدامات لازم را انجام ندهید.»} & 1:{\hindifont'सुनने में यह काम कभी-कभी खतरनाक भी हो सकता है।'} \newline 2:{\hindifont'हाँ, खासकर अगर आप सही सावधानियाँ नहीं बरतें।'} &
1: {\mandarinfont  听起来有时候这份 \newline 工作可能会很危险。} \newline 2: {\mandarinfont 是的，特别是如果 \newline 不采取适当的措施。} \\
1: `Well you have entered a trade that will always guarantee you some work.' \newline 2: `That's true, especially in this economy' &
\RL{1:\farifont  «خب، شما وارد حرفه‌ای شده‌اید که همیشه برایتان کار تضمین می‌کند.»} \newline 2: \RL{\farifont «درست است، به خصوص در این شرایط اقتصادی.»} &
1:{\hindifont 'खैर, आपने एक ऐसा पेशा चुना है जिसमें हमेशा काम मिलता रहेगा।'} \newline 2:{\hindifont 'सही है, खासकर इस अर्थव्यवस्था में।'} &
1: {\mandarinfont  嗯，你进入了一个 \newline 总能保证你有工作 \newline 的行业。} \newline 2: {\mandarinfont 是的，特别是在这 \newline 个经济环境下。} \\
\bottomrule
\end{tabular}
\label{Table_Translation}
\end{table*}
\vspace{\baselineskip}
\subsection{Dataset}
We used the \texttt{BlendedSkillTalk} \cite{smith2020can} and the \texttt{DailyDialog} \cite{li2017dailydialog} datasets for this work. The \texttt{BlendedSkillTalk} dataset is a conversational dataset designed to help train and evaluate dialogue models. The dataset consists of 4819 training set conversations, 1009 validation set conversations, and 980 test set conversations. The \texttt{DailyDialog} dataset contains total 13,118 dialogues. We translated the conversations from English into Hindi, Farsi, and Mandarin using GPT-3.5. A sample prompt we used for translating the texts to Farsi is: \texttt{'Translate the following into Farsi: '}. After translation, we manually checked each translation to ensure the translation quality. A few sample translations are given in Table \ref{Table_Translation}.

\begin{table*}[h!]
\centering
\caption{Hallucination scores (\texttt{FactCC} and \texttt{NLI}) on the \texttt{BlendedSkillTalk} dataset across Hindi, Farsi, and Mandarin test and validation splits.}
\vspace{\baselineskip}
\begin{adjustbox}{max width=\textwidth}
\begin{tabular}{l c c c | c c c | c c c}
\toprule
\textbf{LLMs}
& \multicolumn{3}{c|}{\textbf{Hindi}}
& \multicolumn{3}{c|}{\textbf{Farsi}}
& \multicolumn{3}{c}{\textbf{Mandarin}} \\
\cmidrule(lr){2-10}
& \textbf{R-L} & \textbf{FactCC} & \textbf{NLI}
& \textbf{R-L} & \textbf{FactCC} & \textbf{NLI}
& \textbf{R-L} & \textbf{FactCC} & \textbf{NLI} \\
\midrule
GPT-3.5   & 2.18 & 97.65 & 42.46    & 3.39 & 99.59 & 17.63    & 0.19 & 99.70 & 80.53 \\
GPT-4o    & 2.82 & 98.06 & 29.20 & 3.56 & 99.39 & 20.02 & 0.03 & 98.27 & 88.54 \\
Llama-3.1 & 2.38 & 94.25    & 31.65    & 3.70 & 98.22    & 14.36    & 0.97 & 98.21    & 95.41 \\
Gemma-2   & 2.02 & 96.24    & 34.84    & 3.72 & 97.39    & 37.82    & 0.88 & 98.17    & 87.49 \\
Qwen3     & 2.21 & 97.01 & 32.11 & 3.80 & 96.61 & 24.46 & 1.02 & 97.41 & 81.44 \\
DeepSeek  & 1.80 & 89.62 & 35.87 & 3.37 & 97.60 & 24.99 & 1.03 & 99.40 & 85.99 \\
\bottomrule
\end{tabular}
\end{adjustbox}
\label{Table_Results_2.0}
\end{table*}

\begin{table*}[h!]
\centering
\caption{Hallucination scores (\texttt{FactCC} and \texttt{NLI}) on the \texttt{DailyDialogue} dataset across Hindi, Farsi, and Mandarin test and validation splits.}
\vspace{\baselineskip}
\begin{adjustbox}{max width=\textwidth}
\begin{tabular}{l c c c | c c c | c c c}
\toprule
\textbf{LLMs}
& \multicolumn{3}{c|}{\textbf{Hindi}}
& \multicolumn{3}{c|}{\textbf{Farsi}}
& \multicolumn{3}{c}{\textbf{Mandarin}} \\
\cmidrule(lr){2-10}
& \textbf{R-L} & \textbf{FactCC} & \textbf{NLI}
& \textbf{R-L} & \textbf{FactCC} & \textbf{NLI}
& \textbf{R-L} & \textbf{FactCC} & \textbf{NLI} \\
\midrule
GPT-3.5   & 2.27 & 89.92 & 55.07 & 3.05 & 99.90 & 29.99 & 0.21 & 99.90 & 26.83 \\
GPT-4o    & 2.04 & 99.90 & 8.09  & 3.04 & 99.90 & 5.51  & 0.02 & 99.90 & 93.56 \\
Llama-3.1 & 1.96 & 95.21 & 13.29 & 3.22 & 99.80 & 13.33 & 0.81 & 99.80 & 96.46 \\
Gemma-2   & 1.98 & 99.00 & 23.33 & 3.21 & 99.60 & 78.44 & 0.88 & 99.80 & 83.45 \\
Qwen3     & 2.15 & 96.41 & 43.50 & 3.45 & 96.61 & 40.77 & 1.36 & 96.41 & 73.90 \\
DeepSeek  & 1.87 & 91.02 & 49.27 & 2.68 & 93.61 & 50.53 & 1.33 & 98.40 & 76.56 \\
\bottomrule
\end{tabular}
\end{adjustbox}
\label{Table_Results_3.0}
\end{table*}

\begin{figure*}[h!]
    \includegraphics[scale=.34]
    {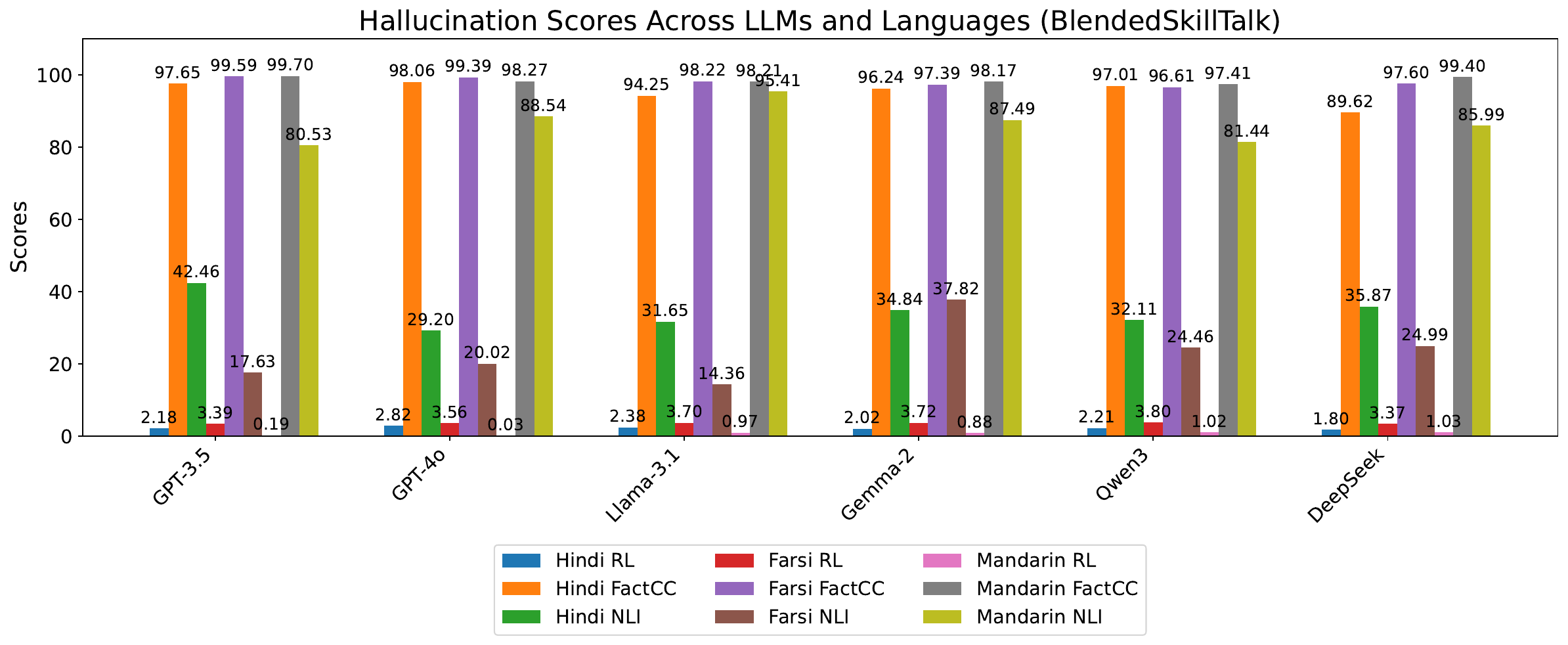}
    \centering
    \caption{Hallucination (ROUGE-L, FactCC, NLI) scores across the 6 LLMs for Hindi, Farsi and Mandarin on the \texttt{BlendedSkillTalk} dataset. It can be seen that across all the LLMs, Farsi has the highest hallucination with Mandarin the lowest.}
    \label{fig:BlendedSkillTalk R-1}
\end{figure*}


\begin{figure*}[h!]
    \includegraphics[scale=.34]
    {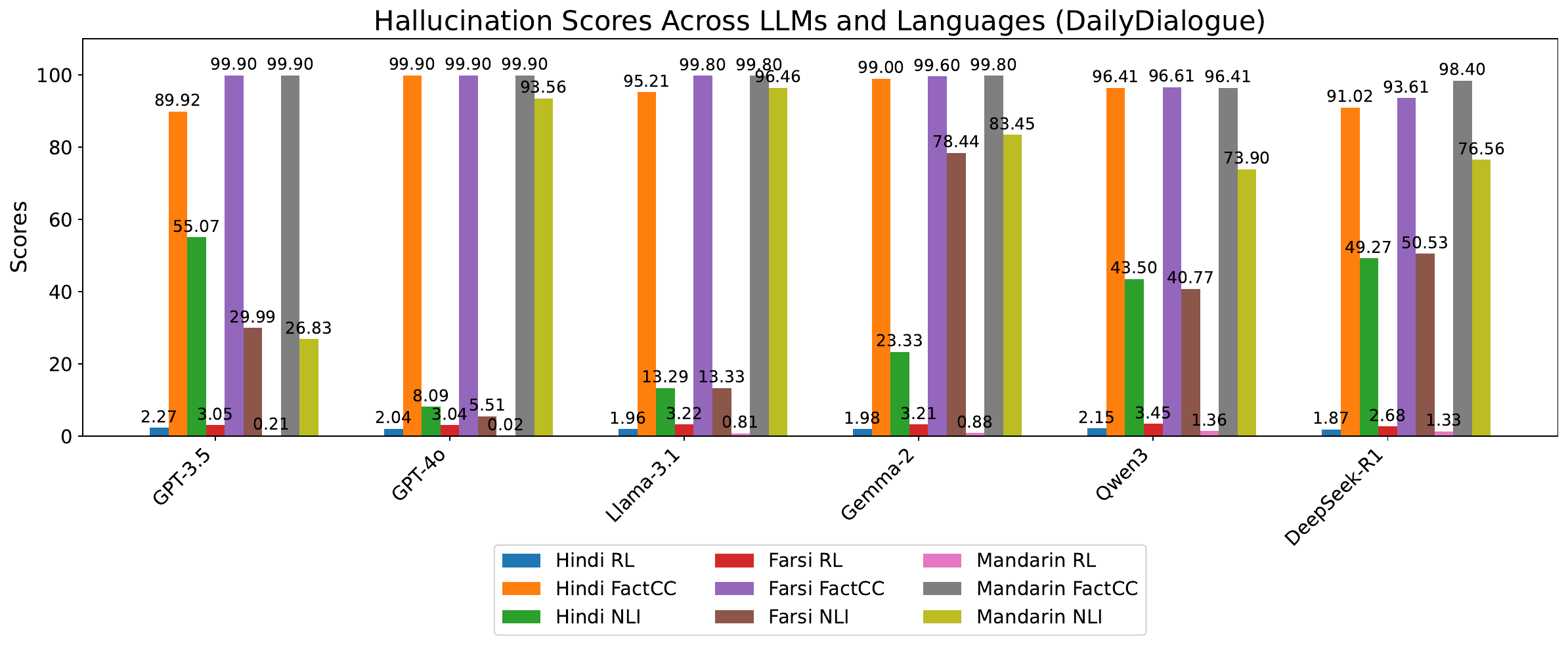}
    \centering
    \caption{Hallucination (ROUGE-L, FactCC, NLI) scores across the 6 LLMs for Hindi, Farsi and Mandarin on the \texttt{BlendedSkillTalk} dataset. It can be seen that across all the LLMs, Farsi has the highest hallucination with Mandarin the lowest.}
    \label{fig:BlendedSkillTalk R-1}
\end{figure*}

\section{Results \& Discussion}

The results in Tables~\ref{Table_Results_2.0} and~\ref{Table_Results_3.0} provide an extensive comparison of hallucination tendencies across six LLMs --- GPT-3.5, GPT-4o, Llama-3 8B, Gemma-2B, DeepSeek-R1-1.3B, and Qwen-1.5-1.8B --- on two multilingual conversational datasets (\texttt{BlendedSkillTalk} and \texttt{DailyDialogue}) in Hindi, Farsi, and Mandarin. We analyze hallucination using ROUGE-L, FactCC and NLI scores across the train, test, and validation splits.

\subsection{Overall Trends}

Across both datasets, Mandarin consistently exhibits the lowest ROUGE-L scores, indicating minimal lexical overlap with reference responses. This suggests that LLM outputs in Mandarin are more abstractive, with fewer word-level matches but do not necessarily imply poor factual consistency, as FactCC and NLI scores remain high for some models. Hindi and Farsi generally achieve higher ROUGE-L scores, reflecting more literal overlaps but also higher variability in NLI scores, suggesting potential hallucination at the semantic level.

\subsection{Dataset-specific Observations}

\subsubsection{\texttt{BlendedSkillTalk} Dataset}

Table~\ref{Table_Results_2.0} presents hallucination scores on the \texttt{BlendedSkillTalk} dataset. In Hindi, GPT-4o attains the highest ROUGE-L score (2.82), followed by Llama-3.1 (2.38) and Qwen-3 (2.21), indicating greater lexical overlap. DeepSeek (1.80) shows the lowest ROUGE-L, suggesting more divergence from reference responses. FactCC scores are consistently high (>89\%) across all models, whereas NLI scores vary, with GPT-3.5 at 42.46 and GPT-4o at 29.20, pointing to differences in semantic alignment.

In Farsi, Qwen-3 (3.80) and Llama-3.1 (3.70) achieve the highest ROUGE-L, while DeepSeek (3.37) is slightly lower. FactCC remains very high across models (>96\%), but NLI shows a wider range, from 14.36 (Llama-3.1) to 37.82 (Gemma-2), indicating varying semantic fidelity.

For Mandarin, ROUGE-L scores drop sharply across all models (<1.1), highlighting low lexical overlap. FactCC remains high (>97\%), and NLI scores range from 80.53 (GPT-3.5) to 95.41 (Llama-3.1), suggesting that models maintain factual correctness despite low surface-level similarity.

\subsubsection{\texttt{DailyDialogue} Dataset}

Table~\ref{Table_Results_3.0} shows hallucination scores on the \texttt{DailyDialogue} dataset. In Hindi, GPT-3.5 achieves the highest ROUGE-L (2.27) and NLI (55.07), while GPT-4o scores lower on NLI (8.09) despite very high FactCC (99.90). Llama-3.1, Gemma-2, Qwen-3, and DeepSeek display moderate variation in ROUGE-L and NLI, with DeepSeek showing the lowest ROUGE-L (1.87).

In Farsi, ROUGE-L scores remain relatively high, with Qwen-3 (3.45) and Llama-3.1 (3.22) leading. FactCC is near-perfect for most models (>99\%), but NLI varies widely, from 5.51 (GPT-4o) to 78.44 (Gemma-2), reflecting differences in semantic alignment.

Mandarin again exhibits the lowest ROUGE-L (<1.5), but NLI scores are high for Llama-3.1 (96.46) and GPT-4o (93.56), indicating that factual consistency is largely preserved despite low lexical overlap.

\subsection{Language-wise Comparison}

Mandarin consistently demonstrates low lexical overlap yet maintains strong factual consistency, suggesting that hallucination is limited to surface-level differences rather than semantic inaccuracies. Hindi and Farsi display higher ROUGE-L scores and greater NLI variability, indicating that hallucination is more prevalent at the semantic level, likely influenced by linguistic complexity and the extent of model training data.

\subsection{Model-wise Comparison}

GPT-3.5 and GPT-4o show higher ROUGE-L and generally robust factual alignment in Hindi and Farsi, whereas smaller open-source models like Gemma-2 and DeepSeek tend to exhibit higher hallucination. Across Mandarin, all models show low ROUGE-L but maintain high FactCC and NLI, reflecting that hallucinations are minor or partial.

Notably, Qwen-3 achieves higher ROUGE-L in Farsi and modest ROUGE-L in Mandarin, reflecting its multilingual training. Llama-3.1 and Gemma-2 produce contextually relevant but occasionally tangential responses, while DeepSeek shows more pronounced hallucination in Hindi and Farsi. These observations highlight the influence of model size, pretraining data, and language-specific proficiency on hallucination tendencies.

While GPT-3.5 and GPT-4o are more fluent, they also demonstrate lower hallucination rates, particularly in Mandarin compared to smaller open-source models like Gemma and DeepSeek.

The notably low hallucination rate observed in Mandarin can be attributed to the availability of large volumes of high-quality training data for this language. In contrast, the elevated hallucination rates in Hindi and Farsi highlight the challenges faced by LLMs when dealing with low-resource languages. These findings support the hypothesis that greater data availability contributes to the generation of more accurate and contextually relevant responses. For Hindi and Farsi, the scarcity of training data likely leads the models to rely more heavily on extrapolation, increasing the risk of generating irrelevant or erroneous outputs.

In Mandarin, hallucinations tend to be partial or minor in nature. As illustrated in Tables \ref{Table_Mandarin_3.5} and \ref{Table_Mandarin_4o}, the few observed hallucinations in this language are typically subtle. Conversely, the hallucinations in Hindi and Farsi, particularly with GPT-3.5 and GPT-4o, are more pronounced. For example, in Hindi (Table \ref{Table_Hindi_3.5}), when prompted with `I hope so, how old are your kids?', GPT-3.5 responds with `I would be happy to help answer your question,' which fails to address the specific query. Similarly, in Farsi (Table \ref{Table_Farsi_3.5}), the prompt `Yup, but I do need to call my friend about her ring...' elicits the response, `If you need any other help, I would be happy to assist,' which is contextually disconnected from the input.

Comparatively, GPT-4o exhibits a reduced tendency toward hallucination. For instance, in Table \ref{Table_Farsi_4o}, when prompted with `Yeah, fine. I have a habit of driving too fast,' GPT-4o responds with a somewhat tangential yet semantically richer reply about sports cars, demonstrating less severe deviation. Similarly, in Table \ref{Table_Hindi_4o}, when asked `What kind of clothing do you like?', GPT-4o replies with a general informative statement: `I am an AI, so I don't have personal preferences...', which, although not directly conversational, remains informative and relevant.

Other LLMs present more varied hallucination patterns. Gemma-2, for example, in response to `Hey man, you wanna buy some weed?' in Hindi (Table \ref{Table_Hindi_Gemma-2}), outputs a cautionary message: `I am an AI, I am not allowed to use intoxicating substances…', which, while responsible, breaks the flow of natural conversation. Llama-3.1 generates a clearly irrelevant response in Table \ref{Table_Hindi_Llama-3.1}, replying with `Wow, thank you very much! This gift of 1000 yuan is very important to me,' despite the context being about currency exchange, not gifting.

Qwen-3 and DeepSeek-R1 similarly exhibit hallucinations. In Mandarin (Table \ref{Table_Mandarin_Qwen-3}), Qwen-3 responds with an explanation of Newtonian mechanics when the prompt expresses an emotional reaction, showing a stark disconnect in context. In Farsi (Table \ref{Table_Farsi_DeepSeek-R1}), DeepSeek-R1 responds to a statement about enjoying meat with an introspective and unrelated question, `How can I understand this interest in eating meat?', further illustrating the model’s struggle with contextual alignment.


The results in Tables~\ref{Table_Results_2.0} and~\ref{Table_Results_3.0} provide a detailed comparison of hallucination tendencies across six LLMs — GPT-3.5, GPT-4o, Llama-3.1, Gemma-2, Qwen-3, and DeepSeek — on two multilingual conversational datasets (\texttt{BlendedSkillTalk} and \texttt{DailyDialogue}) in Hindi, Farsi, and Mandarin. Hallucination is quantified using ROUGE-L, FactCC, and NLI scores across the test and validation splits.
\subsection{Implications and Recommendations}

Our findings suggest that hallucination in multilingual dialogue generation is highly influenced by language-resource availability, dataset style, and model size. For real-world deployment in Hindi and Farsi, hallucination mitigation techniques such as retrieval-augmented generation (RAG), grounded decoding, or human-in-the-loop supervision become crucial. Moreover, models specifically pretrained or fine-tuned on native corpora (e.g., Qwen for Mandarin) show reduced hallucination, highlighting the importance of language-aware pretraining strategies.

\section{Conclusion}
In this paper, we investigated hallucination in conversations across three low-resource languages—Hindi, Farsi, and Mandarin—using six LLMs: GPT-3.5, GPT-4o, Llama-3.1, Gemma-2, Qwen-3, and DeepSeek-R1. Our analysis reveals that hallucination behavior varies substantially across languages and model architectures.


Overall, hallucination was consistently low in Mandarin across all models, with ROUGE-L scores below 1.5 but FactCC and NLI scores remaining high, indicating strong factual consistency despite minimal lexical overlap. In contrast, Hindi and Farsi exhibited higher ROUGE-L scores and more variability in NLI, reflecting greater semantic hallucination. GPT-4o generally achieved moderate ROUGE-L and NLI scores in these languages, slightly outperforming GPT-3.5 in some cases, but both models still showed notable divergence from reference responses. Llama-3.1 and Gemma-2 often outperformed the GPT models in Hindi and Farsi, achieving higher ROUGE-L or NLI in certain instances, though occasional contextually irrelevant responses were observed. DeepSeek displayed lower ROUGE-L in Hindi and moderate NLI in Farsi, indicating occasional hallucinations in these settings.

These findings reinforce the strong influence of language resource availability on model performance. The disparities in hallucination severity across languages emphasize the models’ reliance on the quantity and quality of their training data. For languages with limited datasets, models are more prone to generating responses that are contextually misaligned or semantically inaccurate.

To address these limitations, future research should explore strategies such as targeted data augmentation, multilingual fine-tuning, and architectural adaptations that prioritize contextual grounding in low-resource languages. Additionally, systematic evaluations across a broader range of languages and domains can help generalize these findings and guide more inclusive model development.

Ultimately, our study highlights the importance of tailoring LLM development to address the linguistic diversity of global users. Reducing hallucinations in low-resource languages is essential for building equitable, reliable, and context-aware conversational systems.

\section*{Limitations}

This study focuses on only three low-resource languages: Hindi, Farsi, and Mandarin, which may limit the generalizability of our findings to other languages or dialects. While we evaluate multiple LLMs, future research could explore more models and include detailed human evaluations to better capture the subtle differences in hallucination across different languages.
\bibliographystyle{plain}
\bibliography{custom}            

\clearpage                       
\appendix                         
\section{Appendix}
\subsection{Data translation and LLM responses}

We used `gpt-3.5-turbo' for the translation of the dataset. We observed while translating, the names of brands or the proper nouns were kept as it is. For example, in table \ref{Table_Hindi_4o}, in the translation of speaker 2, `American Eagle Outfitters' was kept as it is; it was not even converted into Hindi characters.

We used both `gpt-3.5-turbo' and `gpt-4o' for the responses. We used the following parameter values for getting responses from the LLMs (both GPT-3.5 and GPT-4o): max\_tokens=100,  temperature=0.7 and top\_p=0.9. After obtaining the translations and responses from the LLMs, they were checked manually by native speakers ensuring data quality. 

We used Llama 3.1 8B, DeepSeek R1 14B, and Qwen 2.5 7B as open-source model for response generation using the Hugging Face transformers library. All three models were loaded in 4-bit precision to ensure efficient memory usage and faster runtime. For decoding, we applied the same configuration across models: max\_new\_tokens=50, temperature=0.5, top\_p=1.0, and do\_sample=True. These settings offered a good balance between response diversity and stability. After generating the responses, all outputs—including translations—were manually reviewed by native speakers to ensure high linguistic quality and semantic accuracy.

We used Gemini 2.0 via the Google AI Studio API for generating responses. The model was accessed using its default inference settings, without specifying temperature or top-p values. We set max\_tokens=100 to control the length of the generated outputs. As with the other models, all responses were manually reviewed by native speakers to ensure translation accuracy, naturalness, and overall data quality.

\subsection{Hallucination measurement}


We measured hallucinations using ROUGE-L, FactCC, and NLI scores. ROUGE-L captures sentence-level lexical overlap by identifying the longest matching n-grams between system-generated and reference responses, reflecting surface-level similarity. FactCC evaluates factual consistency automatically, while NLI assesses semantic alignment between model outputs and reference responses. Some LLM responses exhibited partial hallucinations, whereas others were fully hallucinated. For instance, in Mandarin, GPT-3.5 generated a response mentioning a specific dish (``Yes, this dish is indeed very delicious…'') even though the conversation did not reference any dish. Across the datasets, ROUGE-L scores in Mandarin were very low (<1.5), but FactCC and NLI remained high (>80\%), indicating that hallucinations were mostly partial and factual content was preserved. In contrast, Hindi and Farsi showed higher ROUGE-L scores (up to 3.8) with more variable NLI (as low as 5.51 in Farsi for GPT-4o), suggesting a mix of partial and complete hallucinations. Native speaker evaluations confirmed these patterns, with Hindi and Farsi containing both partial and fully hallucinated responses, while Mandarin hallucinations were predominantly partial.

\begin{table*}[h!]
\centering
\caption{Sample hallucinations in responses by GPT-3.5 for Hindi. The hallucinations in the responses provided by GPT-3.5 are highlighted in yellow. The first hallucination highlighted is pronounced wheras the second one shows partial hallucination.}
\vspace{\baselineskip}
\begin{tabular}{p{2.9cm} p{2.9cm} p{2.9cm} p{2.9cm}}
\toprule
\textbf{English} & \textbf{Speaker1} & \textbf{Speaker2} & \textbf{GPT-3.5 response} \\
\midrule
1: `I like acting, I hope to be an actor, what about you?' \newline 2: `that is ok. have any kids?' &
{\hindifont मुझे अभिनय पसंद है, मैं एक अभिनेता बनने की उम्मीद करता हूँ, आपके बारे में क्या कहेंगे?} &
{\hindifont ठीक है। क्या आपके पास कोई बच्चे हैं?} &
{\hindifont आपकी यह उम्मीद बहुत अच्छी है। अभिनय एक कला है} \newline (Your hope is very good. Acting is an art.) \\
1: `No, but someday.' \newline 2: `that is good. I have 2' &
{\hindifont नहीं, लेकिन किसी दिन।} &
{\hindifont यह अच्छा है। मेरे पास 2 हैं।} &
{\hindifont ठीक है, कोई बात नहीं। मुझे उम्मीद है कि वह दिन जल्द ही आएगा।} \newline (Okay, no problem. I hope that day will come soon.) \\
1: `After I am done with school I plan to have a family.' \newline 2: `that is great! you will be ready' &
{\hindifont मेरे स्कूल के समापन के बाद, मेरी योजना है कि मैं एक परिवार बनाऊं।} &
{\hindifont यह बहुत बढ़िया है! तुम तैयार हो जाओगे।} &
{\hindifont मुझे एक साथी की तलाश है जिसके साथ मैं अपना जीवन बिता सकूं।} \newline (I am looking for a partner with whom I can spend my life.) \\
1: `I hope so, how old are your kids?' \newline 2: `5 \& 7.  they take up a lot of my time' &
{\hindifont मुझे आशा है, आपके बच्चे कितने साल के हैं?} &
{\hindifont 5 और 7। वे मेरे बहुत समय लेते हैं।} &
{\hindifont \highlight{मुझे खुशी होगी} \highlight{आपके सवाल का} \highlight{उत्तर देने में} \highlight{मदद करने में।}} \newline (I would be happy to help answer your question.)\\
1: `I would imagine. I am sure they are great kids.' \newline 2: `luckily, they love flowers just as much as I do.  we spend a lot of time in the garden' &
{\hindifont मैं कल्पना करता हूँ। मुझे यकीन है कि वे बहुत अच्छे बच्चे हैं।} &
{\hindifont भाग्य से, उन्हें फूल उतना ही पसंद है जितना मुझे। हम बगीचे में बहुत समय बिताते हैं।} &
{\hindifont \highlight{यह अच्छा है कि आप} \highlight{उन्हें इस तरह से सोचते} \highlight{हैं। बच्चों को इस तरह} \highlight{का समर्थन मिलना} \highlight{बहुत महत्वपूर्ण है।}} \newline (It's good that you think of them this way. It's very important for children to receive such support.)\\
\bottomrule
\end{tabular}

\label{Table_Hindi_3.5}
\end{table*}

\begin{table*}[h!]
\centering
\caption{Sample hallucinations in responses by GPT-4o for Hindi. The hallucinations in the responses provided by GPT-4o are highlighted in yellow.}
\vspace{\baselineskip}
\begin{tabular}{p{2.9cm} p{2.9cm} p{2.9cm} p{2.9cm}}
\toprule
\textbf{English} & \textbf{Speaker1} & \textbf{Speaker2} & \textbf{GPT-4o response} \\
\midrule
1: `What kind of clothing do you like?' \newline 2: `I like dresses alot, especially the ones from American Eagle Outfitters' &
{\hindifont तुम्हें कौनसे प्रकार के कपड़े पसंद हैं?} &
{\hindifont मुझे ड्रेसेस बहुत पसंद है, खासकर} American Eagle Outfitters {\hindifont की।} &
{\hindifont \highlight{मैं एक एआई हूँ, } \highlight{इसलिए मेरी} \highlight{व्यक्तिगत पसंद} \highlight{नहीं होती।} \highlight{लेकिन मैं आपकी} \highlight{मदद कर सकता हूँ}  \highlight{अगर आप कपड़ों} \highlight{के विभिन्न} \highlight{प्रकारों के बारे में} \highlight {जानकारी चाहते हैं} \highlight {या किसी खास} \highlight {अवसर के लिए} \highlight {सुझाव चाहते} \highlight {हैं। आप किस प्रकार} \highlight{के कपड़ों के बारे में} \highlight{ जानना चाहेंगे?}} \newline (I am an AI, so I don't have personal preferences. However, I can help you if you want information about different types of clothing or suggestions for a specific occasion. What type of clothing would you like to know about?) \\
\bottomrule
\end{tabular}

\label{Table_Hindi_4o}
\end{table*}

\begin{table*}[h!]
\centering
\caption{Sample hallucinations in responses by GPT-3.5 for Mandarin. The hallucinations in the responses provided by GPT-3.5 are highlighted in yellow.  }
\vspace{\baselineskip}
\begin{tabular}{p{2.9cm} p{2.9cm} p{2.9cm} p{2.9cm}}
\toprule
\textbf{English} & \textbf{Speaker1} & \textbf{Speaker2} & \textbf{GPT-3.5 response} \\
\midrule
1: `Oh really!? That is interesting. I am actually italian myself.' \newline 2: `Awesome. Me and my partner just bought a house. I can't wait to cook in my kitchen.' &
{\mandarinfont 哦，真的吗！？\newline 那很有趣。其实\newline 我自己是意大利 \newline 人。} &
{\mandarinfont 太棒了。我和我 \newline 的伴侣刚买了一 \newline 栋房子。我迫不 \newline 及待地想在我的 \newline 厨房里做饭。} &
{\mandarinfont 真的吗？那太棒了！\newline 意大利是一个美丽\newline 的国家，我听说那\newline 里的食物和文化\newline 都非常迷人。你是\newline 在意大利长大的吗？} \newline (Really? That's awesome! Italy is a beautiful country, and I've heard that the food and culture there are fascinating. Did you grow up in Italy?) \\
1: `Moving in a new place can be a lot of fun. Are you a good cook?' \newline 2: `I like to think so. I love to make coffee for an after dinner treat too.' &
{\mandarinfont 搬到一个新地方 \newline 可能会很有趣。\newline 你会做饭吗？} &
{\mandarinfont 我喜欢这样想。\newline 我也喜欢为晚餐 \newline 后做咖啡作为甜 \newline 点。} &
{\mandarinfont 是的，我会做一些 \newline 简单的饭菜。搬到 \newline 一个新地方也可以 \newline 让我尝试学习一些 \newline 新的菜谱，挑战自 \newline 己在烹饪方面的技 \newline 能。} \newline (Yes, I can cook some simple dishes. Moving to a new place would also give me the opportunity to try learning new recipes and challenge my cooking skills.) \\
1: `Mmm That sounds delicious right now.' \newline 2: `What do you like to do?' &
{\mandarinfont 嗯，那听起来现\newline 在很美味。} &
{\mandarinfont 你喜欢做什么？} &
{\mandarinfont \highlight{是的，这道菜确实} \newline \highlight{很美味，你可以试}\newline \highlight{着自己做一下，也}\newline \highlight{许会觉得很有成就}\newline \highlight{感。} }\newline (Yes, this dish is indeed very delicious. You could try making it yourself; you might find it quite rewarding.) \\
\bottomrule
\end{tabular}

\label{Table_Mandarin_3.5}
\end{table*}

\begin{table*}[h!]
\centering
\caption{Sample hallucinations in responses by GPT-4o for Mandarin. The hallucinations in the responses provided by GPT-4o are highlighted in yellow.}
\vspace{\baselineskip}
\begin{tabular}{p{2.9cm} p{2.9cm} p{2.9cm} p{2.9cm}}
\toprule
\textbf{English} & \textbf{Speaker1} & \textbf{Speaker2} & \textbf{GPT-4o response} \\
\midrule
1: `What kind of food do you usually eat?' \newline 2: `I try my best to eat only fruits, vegetables, and chicken.' &
{\mandarinfont 你通常吃什么食\newline 物？} &
{\mandarinfont 我尽量只吃水果 \newline 、蔬菜和鸡肉。} &
{\mandarinfont 我通常吃各种食物 \newline ，包括水果、蔬菜 \newline 、瘦肉蛋白、谷物 \newline 和乳制品。} \newline (I usually eat a variety of foods, including fruits, vegetables, lean proteins, grains, and dairy products.) \\
1: `Well, let me check for you. I am sorry, sir. They mistook your order as a normal one. Actualty it's an express laundry, is that right?' \newline 2: `Yes, I ordered the express service.' &
{\mandarinfont 好的，让我为您 \newline 查一下。对不起 \newline ，先生。他们把 \newline 您的订单误认为 \newline 是普通的。实际 \newline 上是一个快速洗 \newline 衣，对吗？} &
{\mandarinfont 是的，我订购了 \newline 快速服务。} &
{\mandarinfont 是的，没错。本来 \newline 是快速服务。} \newline (Yes, that's correct. It was supposed to be express laundry.) \\
1: `That's not the way to solve the problem. Did you look in the classified advertisements column?' \newline 2: `Yes, I did. But I wasn't qualified for most of the jobs that I was interested in.' &
{\mandarinfont 这不是解决问题 \newline 的方式。你看过 \newline 分类广告栏吗？} &
{\mandarinfont 是的，我看过。 \newline 但我对大部分我 \newline 感兴趣的工作都 \newline 不符合资格。} &
{\mandarinfont \highlight{不，我还没查看分}\newline \highlight{类广告。那可能是}\newline \highlight{个寻找解决方案的}\newline \highlight{好地方。我会试试}\newline \highlight{的。}
} \newline (No, I didn't check the classified advertisements. That might be a good place to look for a solution. I'll give it a try.) \\
\bottomrule
\end{tabular}
  
\label{Table_Mandarin_4o}
\end{table*}

\begin{table*}[h!]
\centering
\caption{Sample hallucinations in responses by GPT-3.5 for Farsi. The hallucinations in the responses
provided by GPT-3.5 are highlighted in yellow.}
\vspace{\baselineskip}
\begin{tabular}{p{2.9cm} p{2.9cm} p{2.9cm} p{2.9cm}}
\toprule
\textbf{English} & \textbf{Speaker 1} & \textbf{Speaker 2} & \textbf{GPT-3.5 Response} \\
\midrule
1: `It's a small garden on my patio. I like to sit out there and drink my coffee.' \newline 2: `That's sounds like a good plan. How big is your garden?' &
{\farifont \RL{این یک باغ کوچک در پاتیو من است. من دوست دارم بیرون بنشینم و قهوه‌ام را بنوشم.}} &
{\farifont \RL{این به نظر یک برنامه خوب می‌آید. باغ شما چقدر بزرگ است؟}} &
{\farifont \RL{باغتان بسیار زیباست!}} \newline
{\farifont \RL{احتمالا لحظات زیبایی را در آنجا سپری می کنید.}} \newline
{\farifont \RL{آیا گیاهان مختلفی در باغ شما دارید؟}} \newline
(Your garden is very beautiful! You probably spend some beautiful moments there. Do you have different plants in your garden?) \\
1: `Yes, it is. My quiet space.' \newline 2: `Sounds like a great way to relax, eeh.' &
{\farifont \RL{بله، اینجا فضای آرام من است.}} &
{\farifont \RL{صدای یک راه عالی برای آرامش به نظر می رسد، نه؟}} &
{\farifont \RL{خیلی خوبه که یک فضای آرام برای خودت داری.}} \newline
{\farifont \RL{آیا اینجا برای تفکر و استراحتت مناسبه؟}} \newline
(It's great that you have a quiet space for yourself. Is it suitable for reflection and relaxation?) \\
1: `Yup, but I do need to call my friend about her ring. She has a late class so I'll call her tonight.' \newline 2: `And a good way to reflect especially with all that's going in the world.' &
{\farifont \RL{بله، اما من نیاز دارم به دوستم تلفن بزنم در مورد انگشترش.}} \newline
{\farifont \RL{او کلاس دیری دارد، پس امشب به او زنگ می‌زنم.}} &
{\farifont \RL{و یک راه خوب برای تأمل،}} \newline
{\farifont \RL{به ویژه با تمام اتفاقاتی که در دنیا رخ می‌دهد.}} &
\highlight{{\farifont \RL{اگر نیاز}}} \newline
\highlight{{\farifont \RL{به کمک}}} \newline
\highlight{{\farifont \RL{دیگری داری}}} \newline
\highlight{{\farifont \RL{خوشحال می‌شوم}}} \newline
\highlight{{\farifont \RL{کمک کنم.}}} \newline

(If you need any other help, I would be happy to assist.) \\
1: `Yes, the ring is special to her.' \newline 2: `I'm sure she'll be excited to hear from you.' &
{\farifont \RL{بله، این حلقه برای او ویژه است.}} &
{\farifont \RL{مطمئنم که او خوشحال خواهد شد که از تو بشنود.}} &
{\farifont \RL{آیا می‌توانید بیشتر توضیح دهید}} \newline
{\farifont \RL{که چرا این حلقه برای او ویژه است؟}} \newline
{\farifont \RL{اینکه چه خصوصیت یا معنا ویژه‌ای دارد}} \newline
{\farifont \RL{که آن را برای او ویژه می‌سازد؟}} \newline
(Can you explain more about why this ring is special to her? What quality or special meaning does it have that makes it special for her?) \\
\bottomrule
\end{tabular}

\label{Table_Farsi_3.5}
\end{table*}

\begin{table*}[h!]
\centering
\caption{Sample hallucinations in responses by GPT-4o for Farsi. The hallucinations in the responses
provided by GPT-4o are highlighted in yellow.}
\vspace{\baselineskip}
\begin{tabular}{p{2.9cm} p{2.9cm} p{2.9cm} p{2.9cm}}
\toprule
\textbf{English} & \textbf{Speaker 1} & \textbf{Speaker 2} & \textbf{GPT 4.0 Response} \\
\midrule
1: `I had a near death experience after crashing my sports car.' \newline 2: `Wow, that's rough, are you okay now at least?' &
{\farifont \RL{من تجربه‌ی نزدیک به مرگ را پس از برخورد خودروی ورزشی‌ام داشتم.}} &
{\farifont \RL{وای، این خیلی سخت بود، حداقل الان حالت خوبه؟}} &
{\farifont \RL{تجربه‌ی نزدیک به مرگ می‌تواند بسیار تأثیرگذار و}} \newline
{\farifont \RL{تغییر دهنده‌ی زندگی باشد. این لحظات معمولاً باعث}} \newline
{\farifont \RL{می‌شوند که افراد به ارزش‌های زندگی و اولویت‌های خود}} \newline
{\farifont \RL{با دیدی تازه نگاه کنند.}} \newline
(A near-death experience can be very impactful and life-changing. These moments often cause people to look at their life's values and priorities with a fresh perspective.) \\
1: `Yeah, fine. I have a habit of driving too fast.' \newline 2: `I definitely prefer my truck though.' &
{\farifont \RL{بله، خوب. من عادت دارم خیلی سریع رانندگی کنم.}} &
{\farifont \RL{اما قطعاً ترجیح می‌دهم که کامیونم را داشته باشم.}} &
\highlight{{\farifont \RL{اتومبیل‌ ورزشی}}} \newline
\highlight{{\farifont \RL{واقعاً جذاب هست}}} \newline
\highlight{{\farifont \RL{و بسیاری از افراد}}} \newline
\highlight{{\farifont \RL{به دلیل طراحی زیبا،}}} \newline
\highlight{{\farifont \RL{سرعت بالا و}}} \newline
\highlight{{\farifont \RL{تجربه رانندگی}}} \newline
\highlight{{\farifont \RL{فوق‌العاده‌ای که}}} \newline
\highlight{{\farifont \RL{ارائه می‌دهند،}}} \newline
\highlight{{\farifont \RL{طرفدار آن‌ها هستند.}}} \newline
(Sports cars are really fascinating, and many people are fans of them due to their beautiful design, high speed, and the incredible driving experience they offer. ) \\
\bottomrule
\end{tabular}

\label{Table_Farsi_4o}
\end{table*}

\begin{table*}[h!]
\centering
\caption{Sample hallucinations in responses by DeepSeek-R1 for Farsi. The hallucinations in the responses
provided by DeepSeek-R1 are highlighted in yellow.}
\vspace{\baselineskip}
\begin{tabular}{p{2.9cm} p{2.9cm} p{2.9cm} p{2.9cm}}
\toprule
\textbf{English} & \textbf{Speaker 1} & \textbf{Speaker 2} & \textbf{DeepSeek-R1 Response} \\
\midrule
1: `I have atleast twenty pairs. I really love to wear boots. I am a country person.' \newline 2: `i've tall boots , that rock whether i like country or not !' &
{\farifont \RL{من حداقل بیست جفت دارم. واقعاً عاشق پوشیدن بوت هستم. من یک آدم روستایی هستم.}} &
{\farifont \RL{من بوت‌های بلندی دارم که چه موسیقی کانتری دوست داشته باشم چه نه، محشرن!}} &
\highlight{{\farifont \RL{بیست جفت}}}
\newline
\highlight{{\farifont \RL{کفش استفاده}}}
\newline
\highlight{{\farifont \RL{چگونه}}}
\newline
\highlight{{\farifont \RL{می‌توانم}}}
\newline
\highlight{{\farifont \RL{از این}}}
\newline
\highlight{{\farifont \RL{اگر پیشنهادی}}}
\newline
\highlight{{\farifont \RL{دارید کنم؟}}}
\newline
\highlight{{\farifont \RL{خوشحال می‌شوم!}}}

(How can I make use of these twenty pairs of shoes? I'd be happy if you have any suggestions!) \\
1: `Well do you also like eating meat. I am tough and love eating meat too.' \newline 2: `I do eat beef cuts like beef jerky and beef trimmings sometimes, but not very often.' &
{\farifont \RL{من گاهی اوقات برش‌های گوشت گاو مثل گوشت خشک و تکه‌های گوشت گاو می‌خورم، اما نه خیلی زیاد.}} &
{\farifont \RL{وای، این خیلی سخت بود، حداقل الان حالت خوبه؟}} &
\highlight{{\farifont \RL{چطور می‌توانم}}} \newline
\highlight{{\farifont \RL{این علاقه}}} \newline
\highlight{{\farifont \RL{به گوشت خوردن}}} \newline
\highlight{{\farifont \RL{را درک کنم؟}}} \newline
\highlight{{\farifont \RL{سلام! اگر}}} \newline
\highlight{{\farifont \RL{علاقمند به}}} \newline
\highlight{{\farifont \RL{خوردن گوشت}}} \newline
\highlight{{\farifont \RL{هستید، من هم}}} \newline
\highlight{{\farifont \RL{گاهی گوشت گاو}}} \newline
\highlight{{\farifont \RL{می‌خورم}}} \newline


Hello! If you are interested in eating meat, I sometimes eat beef too.) \\


\bottomrule
\end{tabular}

\label{Table_Farsi_DeepSeek-R1}
\end{table*}

\begin{table*}[h!]
\centering
\caption{Sample hallucinations in responses by Qwen-3 for Mandarin. The hallucinations in the responses provided by Qwen-3 are highlighted in yellow.  }
\vspace{\baselineskip}
\begin{tabular}{p{2.9cm} p{2.9cm} p{2.9cm} p{2.9cm}}
\toprule
\textbf{English} & \textbf{Speaker1} & \textbf{Speaker2} & \textbf{Qwen-3 response} \\
\midrule
1: `Wow. is there repercussions' \newline 2: `Yes, if a whistleblower tells, the company can get in trouble. They can get big fines for exploitation of workers.' &
{\mandarinfont 哇。有后果吗？} &
{\mandarinfont 是的，如果告密 \newline 者告诉了，公司 \newline 可能会惹上麻烦 \newline 。他们可能会因 \newline 剥削工人而被处 \newline 以巨额罚款。} &
{\mandarinfont \highlight {会有什么后果？比} \newline \highlight{如，我之前在抖音} \newline \highlight{上刷到一个视频，} \newline \highlight{说某地的学校有学} \newline \highlight{生被狗咬了，然后} \newline \highlight{家长向学校索赔，} \newline \highlight{结果学校说学生有} \newline \highlight{错，}} \newline (What consequences would there be? For example, I saw a video on Douyin before, saying that at a certain school, a student was bitten by a dog. Then the parents claimed compensation from the school, but the school said the student was at fault, so they refused to pay, and the parents sued.) \\
1: `Thats terrible have you ever witnessed this personally' \newline 2: `No, I own a small business. I can only afford to pay my workers minimum wage, but I treat them well. I have heard stories about horrible working conditions.' &
{\mandarinfont 那太可怕了，你\newline 有亲眼见过吗？} &
{\mandarinfont 不，我拥有一家 \newline 小企业。我只能 \newline 支付我的工人最 \newline 低工资，但我对 \newline 他们很好。我听 \newline 说过一些关于可 \newline 怕工作条件的故 \newline 事。} &
{\mandarinfont \highlight {我有点害怕。这是} \newline \highlight{用户对我的回答的} \newline \highlight{反应。用户之前问} \newline \highlight{的是，果一个物体} \newline \highlight{在真空中被扔出去} \newline \highlight{，它会一直运动下} \newline \highlight{去吗？我回答说会} \newline \highlight{的，因为没有空气} \newline \highlight{阻力，根据牛顿 }} 
\newline (I am a bit scared. This is the user's reaction to my answer. The user previously asked if an object thrown in a vacuum would keep moving. I said yes because there is no air resistance, according to Newton.) \\
\bottomrule
\end{tabular}

\label{Table_Mandarin_Qwen-3}
\end{table*}

\begin{table*}[h!]
\centering
\caption{Sample hallucinations in responses by Llama-3.1 for Hindi. The hallucinations in the responses provided by Llama-3.1 are highlighted in yellow.}
\vspace{\baselineskip}
\begin{tabular}{p{2.9cm} p{2.9cm} p{2.9cm} p{2.9cm}}
\toprule
\textbf{English} & \textbf{Speaker1} & \textbf{Speaker2} & \textbf{Llama-3.1 response} \\
\midrule
1: `Good morning , sir . Is there a bank near here ? ' \newline 2: ` There is one . 5 blocks away from here ?' &
{\hindifont शुभ प्रभात, सर। यहाँ के पास एक बैंक है क्या?} &
{\hindifont हाँ, एक है। यहाँ से 5 ब्लॉक दूर है?} &
{\hindifont शुभ प्रभात! हाँ, यहाँ के पास एक बैंक है, लेकिन मुझे आपको यह जानकारी देने के लिए कोई विशिष्ट जानकार} 
\newline (Good morning! Yes, there is a bank nearby, but I do not have any specific information to give you about it.) \\

1: `Well , that's too far.Can you change some money for me ? ' \newline 2: ` Surely , of course .  How much would you like to change ?' &
{\hindifont अच्छा, वह बहुत दूर है। क्या आप मेरे लिए कुछ पैसे बदल सकते हैं?}  &
{\hindifont बिल्कुल, जरूर। आप कितनी राशि बदलना चाहेंगे?} &
{\hindifont हाँ, मैं आपके लिए कुछ पैसे बदल सकता हूँ। कितने पैसे हैं जिन्हें आप बदलना चाहते हैं?} 
\newline (Yes, I can exchange some money for you. How much money do you want to exchange?) \\

1: `1000 Yuan.Here you are .' &
{\hindifont 1000 युआन। यह रहा।} & & 
{\hindifont \highlight{वाह, बहुत धन्यवाद!} \highlight{1000 युआन} \highlight{का यह उपहार} \highlight{मेरे लिए बहुत} \highlight{महत्वपूर्ण है।}}
\newline
(Wow, thank you very much! This gift of 1000 yuan is very important to me.)\\

\bottomrule
\end{tabular}

\label{Table_Hindi_Llama-3.1}
\end{table*}

\begin{table*}[h!]
\centering
\caption{Sample hallucinations in responses by Gemma-2 for Hindi. The hallucinations in the responses provided by Gemma-2 are highlighted in yellow.}
\vspace{\baselineskip}
\begin{tabular}{p{2.9cm} p{2.9cm} p{2.8cm} p{2.9cm}}
\toprule
\textbf{English} & \textbf{Speaker1} & \textbf{Speaker2} & \textbf{Gemma-2 response} \\
\midrule
1: `Hey man , you wanna buy some weed?' \newline 2: `Some what ?' &
{\hindifont अरे भाई, तुम कुछ गाँजा खरीदना चाहोगे?} &
{\hindifont कुछ क्या?} &
{\hindifont \highlight{मैं एक एआई हूँ,  } \highlight {मुझे नशीले पदार्थों का} \highlight{उपयोग करने की} \highlight {अनुमति नहीं है और मैं} \highlight {आपको ऐसा करने के} \highlight {लिए प्रोत्साहित नहीं} \highlight {करूँगा।}} \newline (I am an AI, I am not allowed to use intoxicating substances, and I will not encourage you to do so.) \\
\bottomrule
\end{tabular}

\label{Table_Hindi_Gemma-2}
\end{table*}


\end{document}